\documentclass[lettersize,journal]{IEEEtran}
\usepackage{amsmath,amsfonts}
\usepackage{algorithm}
\usepackage{array}
\usepackage[caption=false,font=normalsize,labelfont=sf,textfont=sf]{subfig}
\usepackage{textcomp}
\usepackage{stfloats}
\usepackage{url}
\usepackage{verbatim}
\usepackage{graphicx}
\usepackage{algpseudocode}  
\usepackage{multirow}
\def\BibTeX{{\rm B\kern-.05em{\sc i\kern-.025em b}\kern-.08em
    T\kern-.1667em\lower.7ex\hbox{E}\kern-.125emX}}
\usepackage{balance}
\begin{document}
\title{\LARGE \bf
Few-shot Domain Adaptation for IMU Denoising }

\author{Feiyu Yao$^{\dagger}$, Zongkai Wu$^{\dagger*}$ and Donglin Wang$^{*}$
\thanks{$^{\dagger}$ Contributed equally.  $^{*}$ Corresponding author.      }
\thanks{This project is supported by the Key Program
of Natural Science Foundation of Zhejiang
province, China (Grant No. LZ19F020001) and
the Hangzhou Postdoctoral Research Foundation
(Grant No. 103126582001), Westlake University-Muyuan Joint Research Institute (Grant No. 206006022109)}
\thanks{Z. Wu, Z. Wei and D. Wang are with Machine Intelligence Lab (MiLAB), School of Engineering, Westlake University, Hangzhou 310024, China, and Institute of Advanced Technology, Westlake Institute for Advanced Study, Hangzhou 310024, China.

F. Yao is student with Columbia University and works on this project as visiting students at MiLAB, Westlake University.

E-mail:\{yaofeiyu,wuzongkai,wangdonglin\}@westlake.edu.cn,

feiyu.yao@columbia.edu
}}
\markboth{Journal of \LaTeX\ Class Files,~Vol.~18, No.~9, September~2020}%
{How to Use the IEEEtran \LaTeX \ Templates}

\maketitle

\begin{abstract}
Different application scenarios cause IMU to exhibit different error characteristics which will cause trouble to robot application. Most data processing methods have poor adaptability for different scenarios. To solve this problem, we  propose  a few-shot domain  adaptation method. In this work, a domain adaptation framework is considered for denoising the IMU, a reconstitution loss is designed to improve domain adaptability. In addition, in order to further improve the adaptability in the case of limited data, a few-shot training strategy is adopted. In the experiment, we quantify our method on two datasets (EuRoC and TUM-VI) and two real robots (car and quadruped robot) with three different precision IMUs. According to the experimental results, the adaptability of our framework is verified by t-SNE. In orientation results, our proposed method shows the great denoising performance.
\end{abstract}

\begin{IEEEkeywords}
Few-shot Learning, Domain Adaptation, IMU Denoising.
\end{IEEEkeywords}

\section{Introduction}
\IEEEPARstart{I}{nertial} measurement unit (IMU) plays great importance in robotics. It consists of a gyroscope, sensing angular velocities, and an accelerometer, which measuring linear acceleration signals on three axes in space. Both angular velocity accuracy and acceleration accuracy are crucial for robotics. But nowadays robots are equiped with low-cost IMU, which suffers from serious factor errors, axes misalignments and offsets. This makes low-cost IMU denoising vital for robotics.

\begin{figure}[t]
	\includegraphics[scale=0.6]{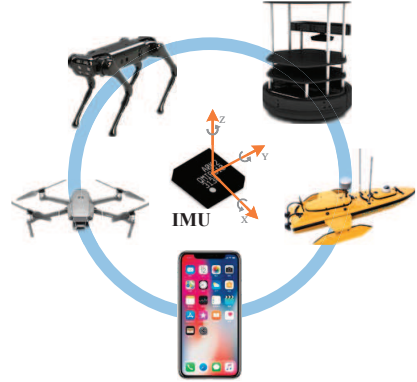}
	\centering
	\caption{IMU is mounted on various devices: from phones and cars in daily life to unmanned aerial vehicles, Quadruped robots, ships in industry.}
	\label{fig1}
\end{figure}

Nowadays, the rapid development in robotics proposes even higher demand for IMU. As shown in Fig. \ref{fig1}, low-cost IMUs are used in various tasks on various platforms. Low-cost IMU has errors from different sources. The error is a mixture of linearity and non-linearity. Common IMU denoising method optimizes designed algorithms for a specific application scenario. From nonlinear filter methods \cite{conventional1}, \cite{wang2018adaptive} and \cite{conventional2} to neural network methods \cite{nn1}, \cite{wu2019ins} and \cite{nn2}, these methods perform well in dealing with IMU errors for specific tasks after optimization with labelled data. However, different robot platforms have different movement patterns, which can seriously influence IMU denoising performance. Also different IMUs have different error characteristics. Thus different IMU on different platform for different task can be seen as different domains. The denoising method mentioned can just perform well in one specific domain. Their optimized parameters can hardly work for other different domains. The huge gap between different tasks in different platform makes it difficult for learned denoisng models to be directly shared among different platforms. 

Few-Shot Learning has been recently leveraged to solve domain adaptation problems. The pursuit is to train a model with the "learning to learn" ability. So the model uses just small amount of training samples to learn and can handle tasks in new domain. A remarkable work in few-shot learning is Model-Agnostic Meta-Learning (MAML) \cite{MAML}. In this work, the parameters are trained to generalize well to new domain tasks with just a small number of gradient steps with a small amount of training samples from that task. To gain this, the sensitivity of the loss functions of new domain tasks with respect to the parameters is maximized in the learning process. MAML is model-agnostic so it can be applied in many learning problems such as classification, regression and reinforcement learning. And \cite{MAML-1}, \cite{MAML-2}, \cite{MAML-3}, \cite{wang2020visual} and \cite{MAML-4} all prove that MAML has great performance and can be used to many models theoretically.


Inspired by this, to solve the generalization problem, we propose IMU denoising method with few-shot domain adaptation ability. The method composes of domain adaptation IMU denoising framework and few-shot learning strategy. The domain adaptation IMU denoising framework models the errors and gets rid of them. It contains Embedding module, Restructor module and Generator module. The 
denoised data is output by Generator module after Embedding module representation. To improve generalization ability of the whole framework, the Restructor along with reconstitution loss is employed for Embedding module to learn better representation in high dimensional space. Besides, to make the framework quickly adapt to the new domain, we propose few-shot learning strategy which enables the framework to further improve performance in the new domain after obtaining a small amount of labelled data.

The main contributions of this work are as followed:

\begin{itemize}
    \item We are the first to take notice of the low-cost IMU denoising generalization problem brought by different IMUs in different application scenarios and locate the problem to the Embedding module.
    \item We propose an IMU denoising method composed by a domain adaptation framework and a corresponding few-shot learning strategy. Proposed IMU denoising method can adapt to a new domain with few labelled data after being trained. 
    \item We implement our proposed IMU denoising method both on open dataset (EuRoC and TUM-VI) and two real robot (car and Quadruped robots) with multiple IMUs. The performance verifies the effectiveness of our method.
\end{itemize}

\section{Task Formulation And Modeling} 

The error characteristics of gyroscope differ among multiple IMUs on multiple robots for different tasks. We define them as different domains. 
Given noisy and biased IMU measurements of angular velocity and acceleration, our domain adaptation method is to denoise angular velocity in multiple domains without further updating any parameter and output accurate orientation estimation after trained.

We consider specific domain tasks $T$ made up by IMU sequences from multiple domains. The measurement of acceleration is represented as $\mathbf{a}_{n}$ and the noisy and biased measurement of gyroscope is expressed as  $\boldsymbol{\omega}_{n}$. The estimated angular velocity is $\hat{\boldsymbol{\omega}}_{n}$. 
The orientation estimation is $\hat{\mathbf{R}}_{n}$.


The orientation estimation is an integration of orientation increments, the process of which can be modeled as following.
\begin{equation}
\hat{\mathbf{R}}_{n}=\hat{\mathbf{R}}_{n-1} \exp \left(\boldsymbol{\hat{\omega}}_{n} d t\right)
\end{equation}
where $\boldsymbol{\omega}_{n} \in \mathbb{R}^{3}$ is the average velocity estimation during $dt$. $\exp (\cdot)$ is the $SO(3)$ exponential map. This model maps the IMU frame to the global frame by successively integrating. Thus the errors can propagate and result in time-varying offsets.

Referring to \cite{ORI}, the measurements of low-cost IMU for calibration is modeled as, 
\begin{equation}
\mathbf{u}_{n}^{\mathrm{IMU}}=\left[\begin{array}{c}
\boldsymbol{\omega}_{n}^{\mathrm{IMU}} \\
\mathbf{a}_{n}^{\mathrm{IMU}}
\end{array}\right]=\mathbf{C}\left[\begin{array}{c}
\boldsymbol{\omega}_{n} \\
\mathbf{a}_{n}
\end{array}\right]+\mathbf{b}_{n}+\boldsymbol{\eta}_{n}
\end{equation}
, in which $\boldsymbol{\eta}_{n} \in \mathbb{R}^{6}$ represents zero-mean, white Gaussian noises and $\mathbf{b}_{n} \in \mathbb{R}^{6}$ are quasi-constant biases.The acceleration $\mathbf{a}_{n}$ without the gravity effect has the following modeling. 

 $\mathbf{a}_{n}=\mathbf{R}_{n-1}^{T}\left(\left(\mathbf{v}_{n}-\mathbf{v}_{n-1}\right) / d t-\mathbf{g}\right) \in \mathbb{R}^{3}$.  Here acceleration is in the IMU frame and $\mathbf{v}_{n} \in \mathbb{R}^{3}$ is IMU velocity in global frame. For the low-cost, consumer grade IMU, the calibration parameter can be approximated by following matrix.

 \begin{equation}
\mathbf{C}=\left[\begin{array}{cc}
\mathbf{S}_{\boldsymbol{\omega}} \mathbf{M}_{\boldsymbol{\omega}} & \mathbf{A} \\
\mathbf{0}_{3 \times 3} & \mathbf{S}_{\mathbf{a}} \mathbf{M}_{\mathbf{a}}
\end{array}\right] 
\end{equation}
. Here $\mathbf{S}_{a}$ and $\mathbf{S}_{\omega}$ are diagonal matrices comprising scaling effects. $\mathbf{M}_{a}$ and $\mathbf{M}_{\omega}$ are lower unitriangular matrices with lower off-diagonal elements corresponding to misalignment small angles.
 A is a fully populated matrix related to the g sensitivity. The scaling effect, misalignment and g sensitivity don't have high impact in low-cost IMU. Thus the model can be changed into increments and the calibration parameter can be estimated by iteration from identity matrix.

Referring to \cite{ORI}, we can model the noise-free angular velocity as:

\begin{equation}\label{model}
\hat{\boldsymbol{\omega}}_{n}=\hat{\mathbf{C}}_{\boldsymbol{\omega}} \boldsymbol{\omega}_{n}^{\mathrm{IMU}}+\boldsymbol{\omega}'_{n}
\end{equation}

Here, the intrinsic parameters $\hat{\mathbf{C}}_{\boldsymbol{\omega}}=\hat{\mathbf{S}}_{\boldsymbol{\omega}} \hat{\mathbf{M}}_{\boldsymbol{\omega}} \in \mathbb{R}^{3 \times 3}$. The gyro bias $\boldsymbol{\omega}'_{n} = \hat{c}_{n} + \hat{b}_{n}$, in which $\hat{c}_{n}$ is a time-varying variable and $\hat{b}_{n}$ is static bias. So we need to compute $\boldsymbol{\omega}'_{n}$ and $\hat{\mathbf{C}}_{\boldsymbol{\omega}}$. Notice that they have the above relations and the IMU won't change in a task. So if one is accurately estimated, the other can be optimized. Thus we apply neural network to estimate $\hat{\omega}_{n}$. For $\hat{\mathbf{C}}_{\boldsymbol{\omega}}$, we initialize it as a unit matrix $I\in \mathbb{R}^{3 \times 3}$ and then optimize it during the training process.

\begin{figure*}[t]
	\includegraphics[scale=0.5]{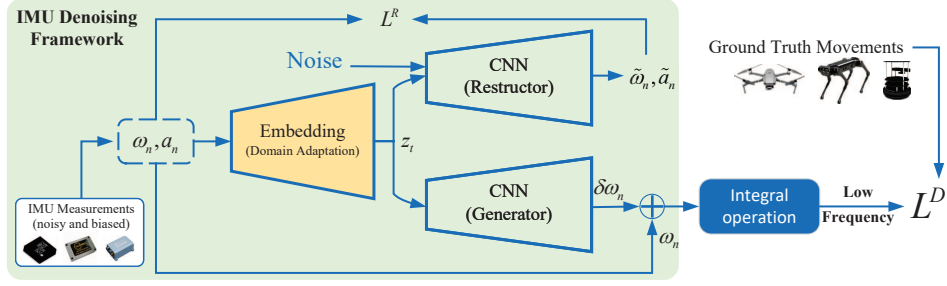}
	\centering
	\caption{Framework of our method, we divide the framework into three modules: (1) an Embedding module; (2) a Restructor to help Embedding; (3) a Generator to denoising. The Embedding module is learned through few-shot learning. The green part represents the framework G. The Embedding module together with Restructor module represents the Restruction process $R$. The Embedding module together with Generator module represents the angular velocity error denoising process $D$.}
	\label{framework}
\end{figure*}

\section{FEW-SHOT DOMAIN ADAPTATION}

In this section, we propose a few-shot domain adaptation IMU denoising method. It consists of a framework and few-shot learning strategy. Firstly, the framework and its key modules are introduced. The IMU denoising generalization problem is located to the Embedding module. Then the few-shot training strategy is specially designed for the Embedding module.

We use the following parameters when describing the training strategy for our framework.

$R$ and $D$:  $R$ represents the reconstruction process and $D$ represents the angular velocity error denoising process.

$G$: $G$ represents the main framework involved in training (The green part in Fig. \ref{framework}). It is composed by the two process $R$ and $D$.

$\theta_{e}$, $\theta_{r}$ and $\theta_{g}$: We use parameters $\theta_{e}$, $\theta_{r}$ and $\theta_{g}$ to represent the Embedding module, Restructor module and Generator module separately. Thus $f$ has parameters $ \theta_{e}$ and $\theta_{r}$ while $h$ has parameters $\theta_{e}$ and $\theta_{g}$.

$\tau_{i}$: $\tau_{i}$ represents a new task.

$\theta_{i}^{\prime}$: $\theta_{i}^{\prime}$ represents the updated model parameters when the new task $\tau_{i}$ comes.

$\omega_{n}$, $a_{n}, \omega^{gt}_{n}$ and $a^{gt}_{n}$: $\omega_{n}$ and $a_{n}$ represent the sequence of noised angular velocity and the sequence of acceleration. $\omega^{gt}_{n}$ and $a^{gt}_{n}$ represent the ground truth angular velocity and acceleration.

\subsection{Framework}

The framework contains mainly three components: Embedding module, Restructor module and Generator module. The Embedding module creates representations where the data from different domains has similar distributions.
The Restructor module reconstructs the sequence for learning better representation. The Generator generates the error estimation in the IMU measurements. The Embedding module and the Restructor module make up the reconstruction process. The Embedding module and the Generator module compose the denoising process. After the integral operation, the IMU measurement compensated by the error estimation will produce the orientation estimation of the robots.

\subsubsection{Module Structure}

The Embedding module structure is Multi-Layer Perception (MLP). The Restructor module and the Generator module are all dilated convolutional neural networks (CNN).

The MLP structure is composed by multiple layers with many following neuron-like processing units in each layer.
 
\begin{equation}
a^{unit}=\phi\left(\sum_{j} w^{unit}_{j} x_{j}+b^{unit}\right)
\end{equation}
 
 where the $x_{j}$ are the inputs to the unit, the $w^{unit}_{j}$ are the weights of one unit, $b$ is the bias of one unit, $\phi$ is the nonlinear activation function, and $a$ is the unit output. The unit output from the former layer will become the input of the next layer.

\begin{figure}[t]
	\includegraphics[scale=0.33]{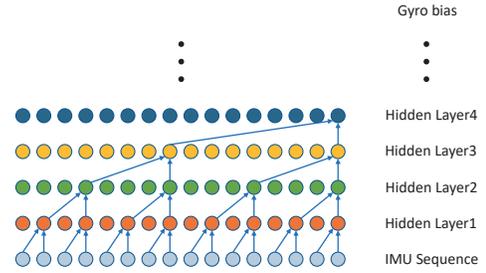}
	\centering
	\caption{Restructor and Generator module structure sketch map}
	\label{DCNN}
\end{figure}

Inspired by \cite{DCNN} and \cite{ORI}, we employ the dilated convolution neural network structure in Generator module to achieve the gyro bias term $\boldsymbol{\omega}'_{n}$ in \eqref{model}:
 
\begin{equation}
\boldsymbol{\omega}'_{n}=f\left(\mathbf{u}_{n-N}^{\mathrm{IMU}}, \ldots, \mathbf{u}_{n}^{\mathrm{IMU}}\right),
\end{equation}
where $\mathbf{u}_{n}^{\mathrm{IMU}}=\left[\begin{array}{l}\boldsymbol{\omega}_{n}^{\mathrm{IMU}} \\ \mathbf{a}_{n}^{\mathrm{IMU}}\end{array}\right]$. The N represents the length of local window of previous measurements, which is the base of correction for current state. As Fig. \ref{DCNN} represents, Generator module $R(\cdot)$ can approximate gyro bias term with five layer dilated convolution neural network structure. The dilation gaps for five layers are 1, 4, 16, 64, 1 separately and the kernel dimensions for each layer are all 7. The gyro bias term will then be brought in \eqref{model} to achieve the denoised angular velocity. Thus it can correct the data smoothly and bring multi-scale time information.

The Restructor module has similar five layer dilation convolution neural network structure except the output dimension. There will be some noise added to the Embedded IMU data before Restructor operations and the output of Restructor is expected to be as similiar as original angular velocity and acceleration measurements.

\begin{figure}[t]
	\includegraphics[scale=0.35]{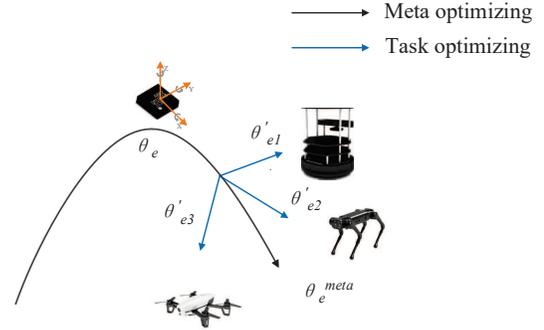}
	\centering
	\caption{Diagram of our few-shot training strategy, which optimizes the parameter $\theta_{e}$ using data from different robots.}
	\label{figure3}
\end{figure}

\subsubsection{Loss function}

For the reconstruction process, the output of the process is expected to greatest extent be restored to the original IMU sequence. Thus we use Mean Square Loss to build the loss function. 
\begin{equation}
L^{R}=E\left\|[\hat{\omega}_n, \hat{a}_n]-[\omega_n,a_n]\right\|_{2}^{2}
\end{equation}

\begin{equation}
[\hat{\omega}, \hat{a}]=R\left(\omega_n, a_n, \theta_r\right)
\end{equation}
, in which $\omega$ is the original data. 

For the denoising process, referring to \cite{ORI}, it is not suitable to use mean-square error since tracking system performance relates to frequency. Thus we apply rotation matrix to build loss function:

\begin{equation}
\emph{L}^{D}_{i}=\sum_{i} \rho\left(\log \left(\delta \mathbf{R}^{gt}_{i, i+j} \delta \hat{\mathbf{R}}_{i, i+j}^{T}\right)\right)
\end{equation} 
in which logarithm map SO(3) is $log(\cdot)$ and Huber loss $\rho(\cdot)$ are contained. The The $\delta \mathbf{R}_{i, i+j}$ is defined as:

\begin{equation}
\delta \mathbf{R}_{i, i+j}=\mathbf{R}_{i}^{T} \mathbf{R}_{i+j}=\prod_{k=i}^{i+j-1} \exp \left(\boldsymbol{\omega}_{k}\right)
\end{equation}

\begin{equation}
\delta \mathbf{R}^{gt}_{i, i+j}=\prod_{k=i}^{i+j-1} \exp \left(\boldsymbol{\omega^{gt}}_{k}\right)
\end{equation}

Since the IMU working frequence tends to higher than the tracking system working frequency, the parameter $j$ is used to reduce the frequency of IMU. 
The loss with different parameter $j$ can be added to achieve the final loss in order to gain better performance.

\subsection{Few-shot Training Strategy}


In real life, labelling data is quite time-consuming and laborious. For better representation with limited data, a few-shot training strategy is applied to learn the domain invariant representation in Embedding module. Referring to Model-Agnostic Meta-Learning (MAML) \cite{MAML} structure in few-shot tasks, we apply meta-training methods to train Embedding module suitable for multiple domains. Fig. \ref{figure3} shows the training method especially for Embedding module.

To construct few-shot learning method, we treat tasks with different application scenarios as meta-tasks $\tau^{d}_{i}$. In each meta-task, the first part of the sequence will be seen as the support set $\tau_{s}$ and the remaining part will be seen as the query set $\tau_{q}$.

\begin{algorithm}[t]  
\caption{few-shot learning strategy} 
\label{alg::conjugateGradient}  
\begin{algorithmic}[1]  
\Require
Sequence set  $T$ containing different domain sequences $\tau^{d}_{1},\tau^{d}_{2},...,\tau^{d}_{n}$, each with IMU data and labeled data.$\theta_{e}$, $\theta_{r}$, $\theta_{g}$: random initialized model parameters.$\alpha$ and $\beta$: hyper-parameters;
\Ensure
Meta-learned Embedding module parameter $\theta_{e}^{meta}$, Restructor module $\theta_{r}$ and Generator module $\theta_{g}$;
\While {not done}
\State Sample batch sequences set $\tau$ from $T$
\State Partition $\tau$ into support set $\tau_{s}$ and query set $\tau_{q}$.
            \ForAll {sequence {$\tau_{si}$} in {$\tau_s$}}:  
                \State  $\theta_{ei}^{\prime}=\theta_{e}-\alpha \nabla_{\theta_{e}}\emph{L}_{\tau_{si}}\left(G_{\theta_{ei}^{\prime},{\theta_{g},\theta_{r}}} \right)$;
            \EndFor 
            \State $\theta_{e,g,r}=\theta_{e,g,r}-\beta \nabla_{\theta_{e,g,r}} \sum_{\tau_{qi} \sim \tau_{q}}  \emph{L}_{\tau_{qi}}(G_{\theta_{ei}^{\prime},{\theta_{g},\theta_{r}}})$;
            \EndWhile
            \State  $[\theta_{e}^{meta},\theta_g,\theta_r]=\theta_{e,g,r}$
    \end{algorithmic}  
\end{algorithm}



The target of training is to obtain the meta parameters $\theta_{e}^{meta}$ by few-shot training and two parameters $\theta_{r}$ and $\theta_{g}$ by normal gradient descent. 
Few-shot training phase is made up by two phases: Task optimizing phase and Meta optimizing phase. Task optimizing phase employs support data to get a set of trained $\theta'_{ei}$, each $\theta'_{ei}$ is suitable for each meta task. Then Meta optimizing phase utilizes the set of trained $\theta'_{ei}$ together with query data to achieve the final meta-trained $\theta'_{ei}$. The data from different domain has similar distribution in Embedding module thus $\theta_{r}$ and $\theta_{g}$ is trained by support data and query data with normal gradient descent. 


In Task optimizing phase, for each task $\tau_{si}$, the Embedding module parameters $ \theta_{e}$ update into $\theta_{e}^{\prime}$. The parameter $\theta_{e}$ is updated by the following gradient descent for several times.

\begin{equation}
\theta_{ei}^{\prime}=\theta_{e}-\alpha \nabla_{\theta_{e}}\emph{L}_{\tau_{si}}\left(G_{\theta_{ei}^{\prime},{\theta_{g},\theta_{r}}} \right),
\end{equation}
in which $\theta_{ei}^{\prime}$ is the updated model parameters in task $\tau_{si}$. $\alpha$ is the fixed hyper-parameter.
$\emph{L}_{\tau_{1i}}\left(G_{\theta_{ei}^{\prime},{\theta_{g},\theta_{r}}}\right)$ has two kinds of loss from reconstruction process and denoising process separately:

\begin{equation}
\emph{L}_{\tau_{si}}=\emph{L}^{R}_{\tau_{si}} +\gamma \emph{L}^{D}_{\tau_{si}},
\end{equation}
in which $\gamma$ is hyper-parameter. For specific task $\tau_{si}$, $\emph{L}^{R}$ is the reconstitution loss in reconstruction process.
$\emph{L}^{D}$ is the loss in the denoising process with multiple frequencies.




In Meta optimizing phase: The model employs the rest data pairs, which are $\omega_{ n}$, $a_{n}$ and $\tilde \omega_{n}$. The model parameters $\theta_{e}$ is updated also through gradient descent, which means:

\begin{equation}
\theta_{e,g,r}=\theta_{e,g,r}-\beta \nabla_{\theta_{e,g,r}} \sum_{\tau_{qi} \sim \tau_{q}}  \emph{L}_{\tau_{qi}}(G_{\theta_{ei}^{\prime},{\theta_{g},\theta_{r}}})
\end{equation}

Here $\theta_{ei}^{\prime}$ are the updated model parameters mentioned in the adaptation optimizing phase and $\beta$ is a fixed hyper-parameter. $\emph{L}_{\tau_{qi}}(G))$ is  the loss in sequence set $\tau_{q}$.
$\theta_{e}$ is updated through meta-training method while $\theta_{r}$ and $\theta_{g}$ are updated through gradient descent method.

After being updated through all several training phases, the final updated model parameters is the meta-trained model parameter $\theta_{meta}$. The whole training algorithm is outlined in Algorithm 1.

\section{Experiment Results}
\subsection{Experiment Setup}
To illustrate the priority of our method, we apply it in public datasets: $\emph{EuRoC}$ \cite{euroc} and $\emph{TUM-VI}$ \cite{tumvi}:

\begin{itemize}
\item $\emph{EuRoC}$: The data comes from a micro aerial vehicle (MAV) equipped with not calibrated ADIS16448 IMU.
This dataset is composed by eleven sequences: MH 01 easy, MH 02 easy, MH 03 medium, MH 04 difficult, MH 05 difficult, V1 01 easy, V1 02 medium, V1 03 difficult, V2 01 easy, V2 02 medium and V2 03 difficult. MH is for industrial machine hall. V1 and V2 is for vicon room but V2 is created more than three months later. Easy, medium, hard mark different flight tasks.
\item $\emph{TUM-VI}$: The data comes from a hand-held device equipped with calibrated BMI160 IMU in different places and different motion modes. The ground truth is made by motion capture system but it only exists in a few sequences. So we use all six sequences with ground truth: Room 1-6.

\end{itemize}
In open datasets experiments, each task is seen as meta task. We set MH 01 easy, MH 02 easy, MH 03 medium, MH 05 difficult, V1 02 medium, V2 01 easy, V2 03 difficult from EuRoC, and room1, room3, room5 from TUM-VI as meta-training sets and set MH 04 difficult, V1 01 easy, V1 03 difficult, V2 02 medium, room2, room4 and room6 as meta-testing sets. In each meta task, data is divided into two parts for learning and validation. Here learning part is for optimizing parameters and validation part is for evaluating bias. Thus the number of tasks from different domains in Task optimizing phase is 10. The support set sequence length is about 60 seconds and the query set is the remaining part of the sequence.

\subsection{Compared Methods}
We compare the following approaches:

\begin{itemize}

    \item GT: This is ground truth angular velocities.
    \item DIGDL: This method \cite{ORI} beats top-ranked IMU denoising algorithms and visual-inertial odometry systems.
    Its training, validation and test sets are the same with our approach.
    
    \item FSDA: This is our framework without few-shot learning strategy for Embedding module.
    
    \item FSDA-F: This is the proposed method with few-shot learning strategy.
    
\end{itemize}

\subsection{Evaluation Metrics}
We evaluate methods by Root Mean Squared Error (RMSE):

\begin{equation}
    \operatorname{RMSE}=\sqrt {\frac{1}{n}\sum\limits_{i = 1}^n {{{\left( \hat {\theta}_i - \theta_i \right)}^2}} } 
\end{equation} 

Here n is the length of the test sequence. $\hat {\theta}_i$ is the estimation  of corresponding orientation angle in $i$ time step. $\theta_i$ is the ground-truth of corresponding orientation orientation angle in $i$ time step. These angles are calculated by integral operation.

\subsection{Results}
TABLE \ref{table} collects the RMSE of DIGDL, FSDA and FSDA-F in all test sequences in three orientation direction. Fig. \ref{resultcurvea} and Fig. \ref{resultcurveb} represent the whole orientation estimation and orientation error for DIGDL, FSDA and FSDA-F in room4 sequence and v1 01 easy sequence. 
In order to verify the effectiveness of our proposed framework and the few-shot learning strategy respectively, we separate the analyse of the result in to two parts.

\subsubsection{proposed framework}

To demonstrate the importance of our proposed method's framework, we compare the orientation errors between DIGDL and FSDA. 
We choose RMSE as our evaluation metric. Also the orientation estimations and orientation errors are visualized for further analysis.

  From TABLE \ref{table}, our proposed method performs $36.21\%$ and $19.11\%$ better than DIGDL respectively on EuRoC and TUM-VI in average RMSE, with lower RMSE performance in all test sequences. 

\begin{figure} [t!]
	\centering
	\subfloat[\label{fig:a}]{
		\includegraphics[scale=0.3]{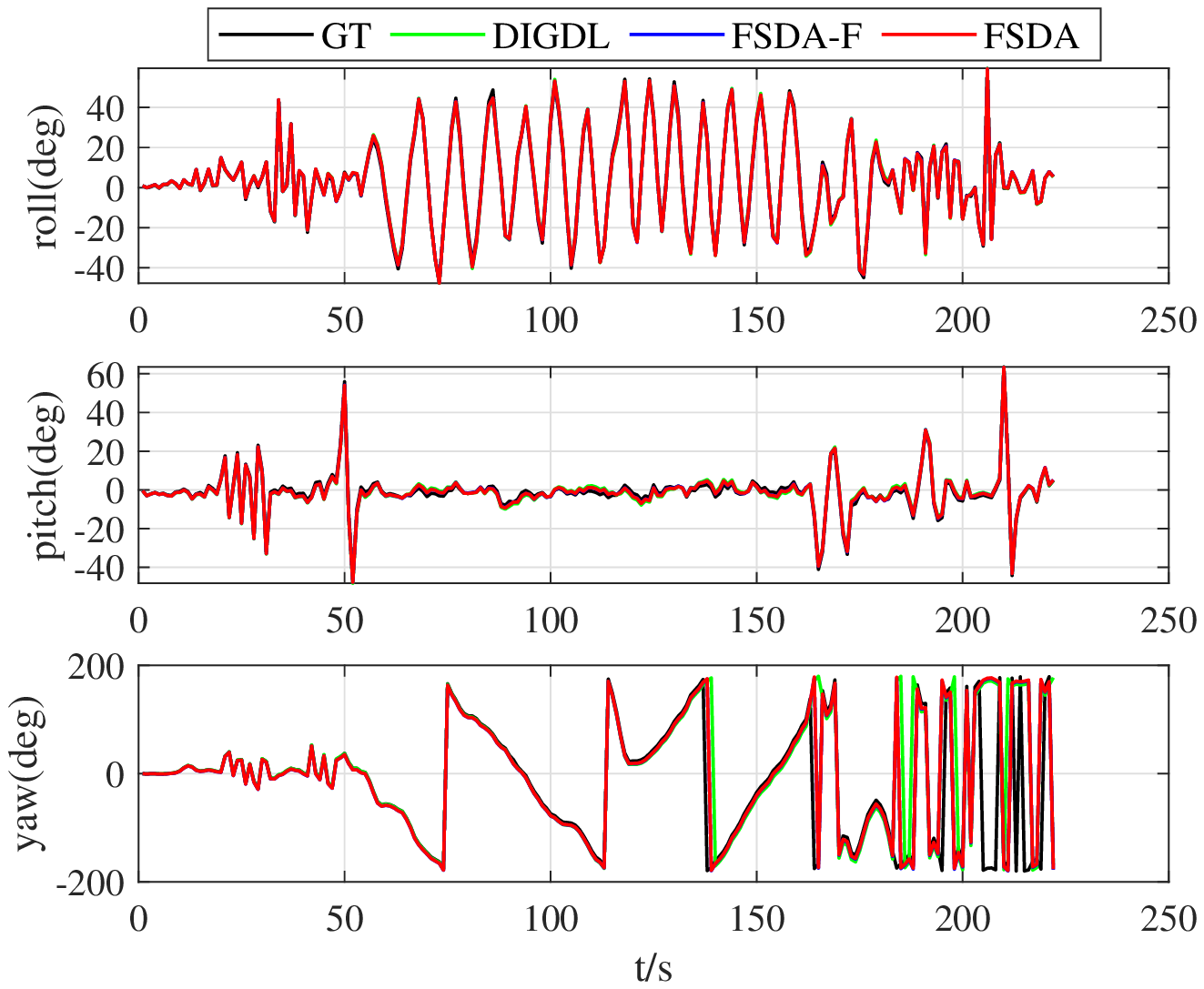}}
	\subfloat[\label{fig:b}]{
		\includegraphics[scale=0.3]{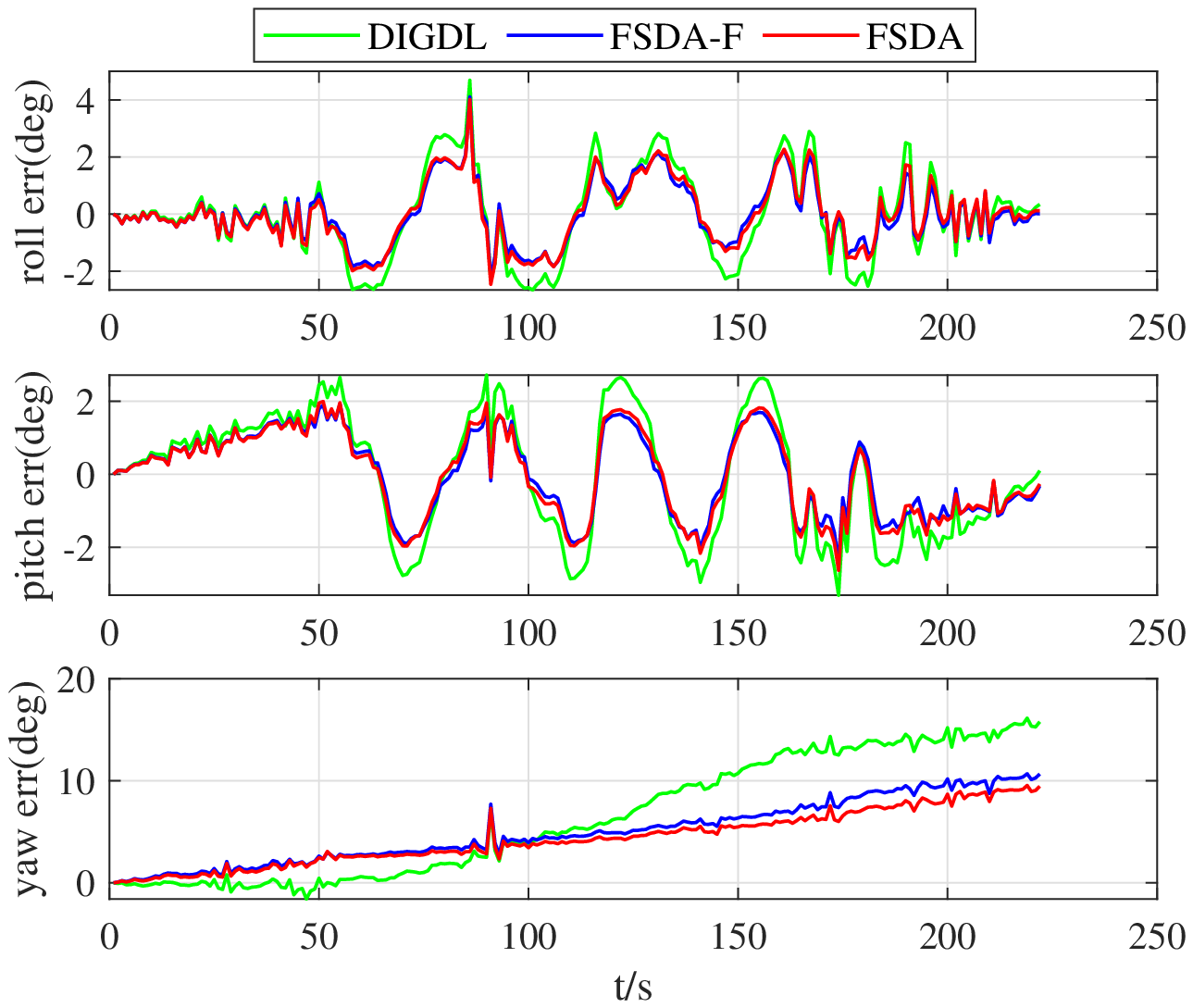}}
	\caption{ a) Orientation estimates; b) Orientation errors on the test sequence room4 for different methods.}
	\label{resultcurvea} 
\end{figure}

\begin{figure} [t!]
	\centering
	\subfloat[\label{fig:a}]{
		\includegraphics[scale=0.3]{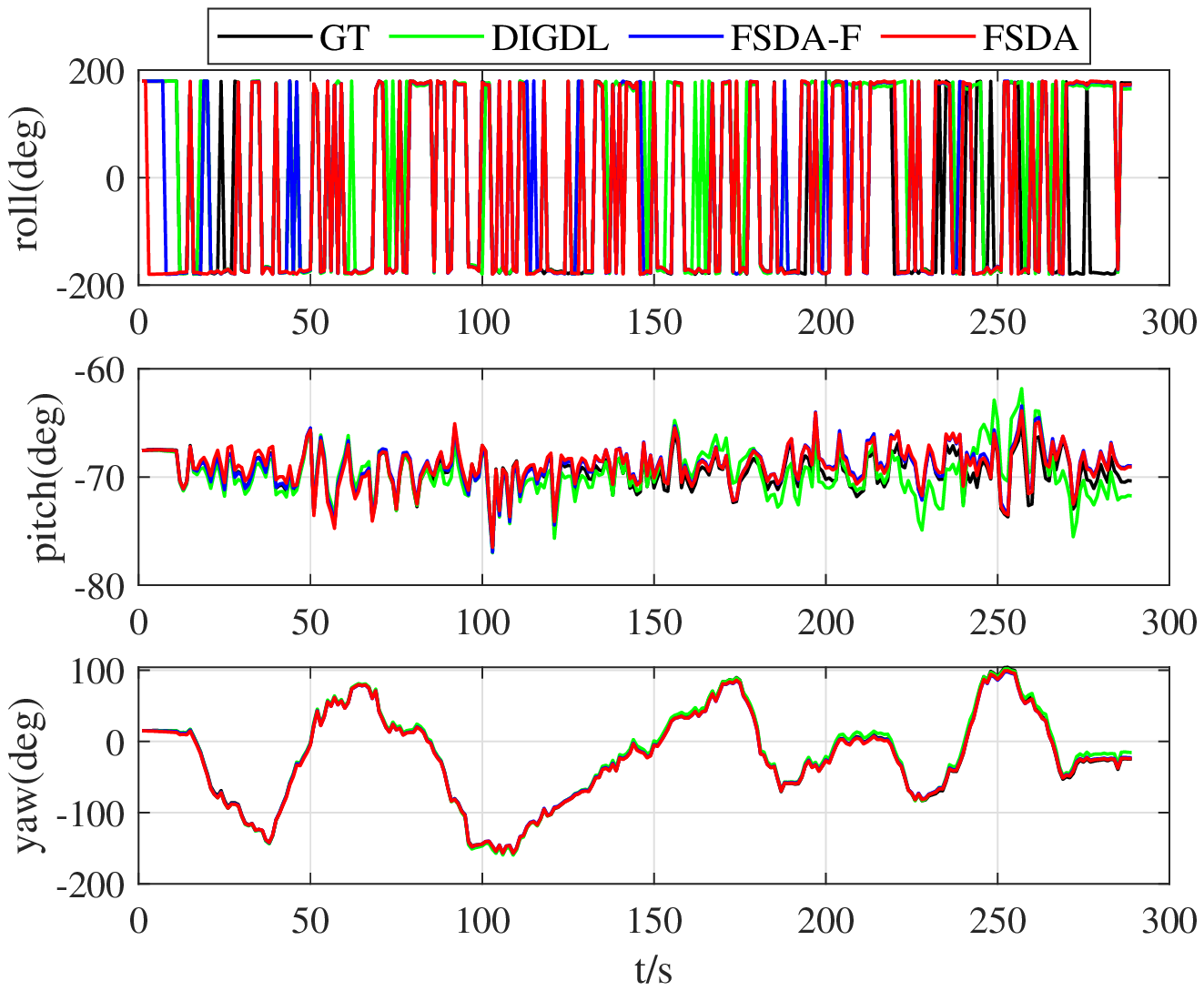}}
	\subfloat[\label{fig:b}]{
		\includegraphics[scale=0.3]{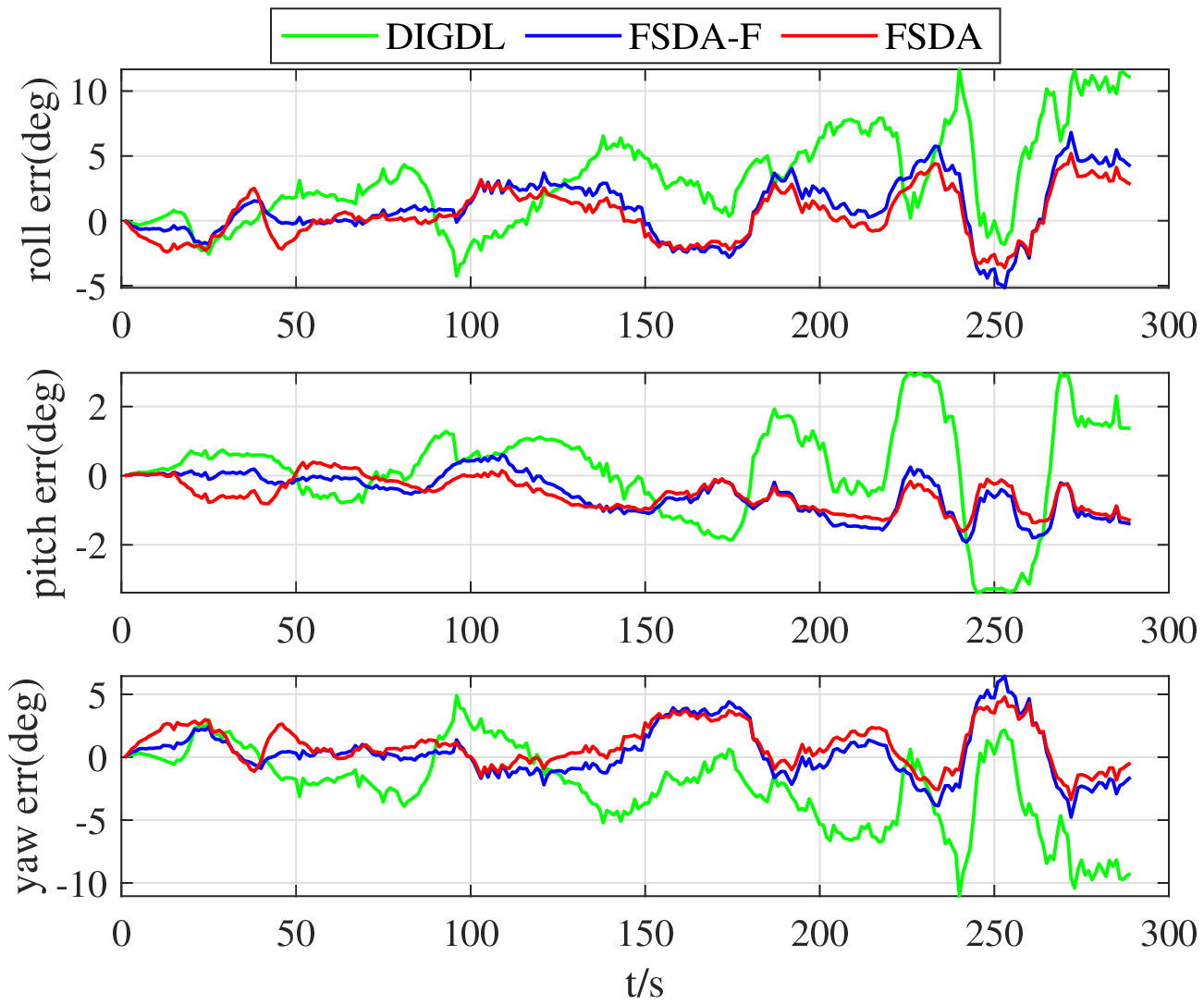}}
	\caption{ a) Orientation estimates; b) Orientation errors on the test sequence V1 01 easy for different methods.}
	\label{resultcurveb} 
\end{figure}

As can be seen from the Fig. \ref{resultcurvea} and Fig. \ref{resultcurveb}, as time goes, the orientation error of DIGDL becomes larger. This can be more serious in the case of large jilters. While our proposed method has good performance even in large jilters. 

This experiment reflects that the proposed framework is effective in low-cost IMU denoising few-shot domain adaptation.

\subsubsection{ proposed few-shot learning strategy}

Here, we take the FSDA denoising performance as baseline to see the performance improvement of the few-shot learning strategy. As shown in TABLE \ref{table}, the FSDA-F has better performance than FSDA in most test sequences.

For the few cases where FSDA performance better in several orientation such as Room4 sequence, we compare the estimation error in the whole test sequence. As can be seen in Fig. \ref{resultcurvea}, the performance of FSDA and FSDA-F are almost the same in the orientation estimations in roll and pitch direction. But in the direction yaw with high variance and large jilters, we can see that the FSDA-F has obvious greater performance in period with high variance. This can also be verified in Fig. \ref{resultcurveb}. This shows that few-shot learning strategy can really improve the denoising performance of low-cost IMU when facing multi-domain tasks.

\begin{table*}[t]
\centering
\caption{Root Mean Squared Error (RMSE) in terms of orientation(roll, pitch and yaw) in degree on the test sequence.}
\renewcommand{\arraystretch}{1.3}
\begin{tabular}{c|c|ccc|ccc|ccc}
\hline
\multirow{2}{*}{Dataset} & \multirow{2}{*}{Sequence} & \multicolumn{3}{c|}{Roll}                               & \multicolumn{3}{c|}{Pitch}                             & \multicolumn{3}{c}{Yaw}                              \\ \cline{3-11}
                         &                           & DIGDL          & FSDA         & FSDA-F             & DIGDL          & FSDA         & FSDA-F             & DIGDL         & FSDA        & FSDA-F             \\ \hline
\multirow{4}{*}{EuRoC}   & MH 04 difficult           & 6.3545          & 4.0661            & \textbf{2.8777}  & 1.4515           & 0.7197 & \textbf{0.5822}          & 5.8232         & 3.3024           & \textbf{2.2552} \\ \cline{2-11} 
                         & V2 02 medium              & 15.2573         & \textbf{13.5855} & 14.8446         & 3.4115             & 3.1405           & \textbf{3.0593}   & 14.6063         & \textbf{12.6688} & 13.9353          \\ \cline{2-11} 
                         & V1 03 difficult           & 15.5216           & 13.0684   & \textbf{5.8844}          & 0.7688 & 0.8006          & \textbf{0.5490}          &15.5827          & 13.0634 & \textbf{6.1500}          \\ \cline{2-11} 
                         & V1 01 easy                & 4.7786          & 2.4963            & \textbf{1.9613}  & 1.3872           & 0.7834          & \textbf{0.7083} & 4.0019          & 2.0675          & \textbf{1.8942}           \\ \hline
\multirow{3}{*}{TUM-VI}  & Room2                     & 1.4942 & 1.4272            & \textbf{1.4066}            & 1.4109           & 1.3223 & \textbf{1.3163}          & 6.1720 & 7.7549          & \textbf{5.8225}          \\ \cline{2-11} 
                         & Room4                     & 1.5116           & \textbf{1.0619}   & 1.1100           & 1.6430  & \textbf{1.1209}          & 1.1762          & 8.6927         & 5.8566          & \textbf{5.0745}  \\ \cline{2-11} 
                         & Room6                     & 1.2296           & 1.0577  & \textbf{1.0124}         & 1.5086           & 1.3003           & \textbf{1.2401}  & 6.5248          & 7.6963          & \textbf{5.6873}  
                         \\ \hline
\end{tabular}

\label{table}
\end{table*}





\subsection{Execution on real robots}
We executed our few-shot domain adaptation method in two real robots and three different types of IMUs.

\begin{figure}[t]
	\includegraphics[scale=0.5]{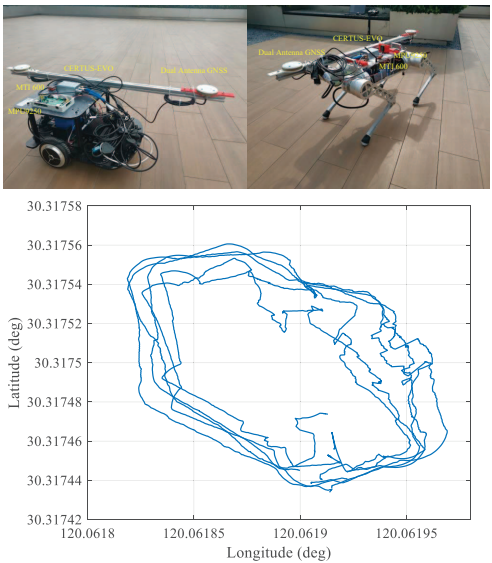}
	\centering
	\caption{Two robots (Car on top left and quadruped robot on the top right), trajectories ( on the down left) and three-axis angular velocity from gyroscopes}
	\label{robot}
\end{figure}

\subsubsection{IMU Setup}

We choose three IMUs with different precision. From high precision to low precision, they are CERTUS-EVO from ADVANCED NAVIGATION, MTI 600 series from XSENS and MPU-9250 from TDK InvenSense:

\begin{itemize}
\item The CERTUS-EVO has the gyroscope with $0.2 ^{\circ}/hr$ bias instability, $6^{\circ} /hr/ \sqrt[]{Hz}$ noise density and $<0.03 \% $ non-linearity, and accelerometer with $8 \mu g$ bias instability, $2 \mu g/\sqrt[]{Hz}$ noise density and $<0.05\%$ non-linearity.

\item The MTI 600 series has the gyroscope with $8 ^{\circ}/hr$ bias instability, $0.007^{\circ} /s/ \sqrt[]{Hz}$ noise density and $0.1 \% FS $ non-linearity, and accelerometer with $15 \mu g$ bias instability, $60 \mu g/\sqrt[]{Hz}$ noise density and $0.1\% FS$ non-linearity.

\item The MPU-9250 has the gyroscope with $0.01^{\circ} /s/ \sqrt[]{Hz}$ noise density and $\pm 0.1 \% $ non-linearity, and accelerometer with $300 \mu g/\sqrt[]{Hz}$ noise density and $\pm 0.5\%$ non-linearity.

\end{itemize}

Since the CERTUS-EVO has dual antenna system, the high-precision orientation sequence from the CERTUS-EVO  can be seen as ground truth.
\subsubsection{Robot Platform}
\begin{figure}[t]
	\includegraphics[scale=0.5]{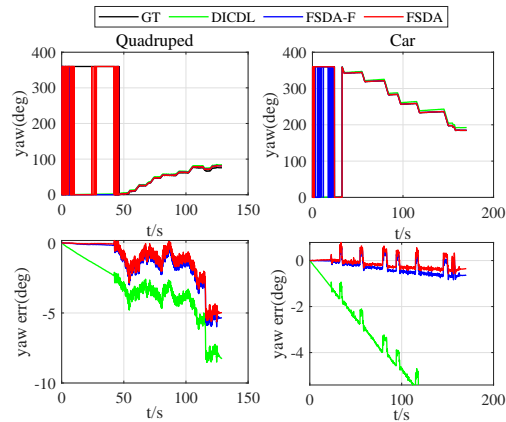}
	\centering
	\caption{The comparison of the orientation estimation and orientation estimation errors of different methods on real robots}
	\label{real-robot}
\end{figure}

Two robot platforms are car and quadruped, which are set up as shown in Fig. \ref{real-robot}. We use all the public datasets and some tasks of car and quadruped robot as the meta-training task. The remaining tasks of car and quadruped robot are set to be meta-testing task. The domain adaptation ability of the Embedding module trained in this experiment is shown in the t-SNE figure Fig.\ref{tsne}.

In implementation, the car robot drives six times and each time it drove in a circle. The quadruped robot walked twice and each time it's trajectory was 3/4 lap. 
Due to page limitation, we choose the yaw angle that is most likely to accumulate errors to show the performance of our method.
The performance is shown in Fig. \ref{real-robot}. As can be seen, our few-shot domain adaptation method works well on real robots and IMUs. Obviously, the divergence of the DIGDL is serious, which means DIGDL has no adaptability in sequence from different domains. Our FSDA framework drastically reduces the error. With the few-shot learning strategy, the performance is further improved.

\subsection{Interpreting the Embedding Module}
 
t-distributed stochastic neighbor Embedding (t-SNE) \cite{t-sne} is a method which can visualize the high-dimensional data by mapping it into a two dimensional space. The mapping is based on the principle that similar objects in high dimensional space are modeled by nearby points in two-dimensional space and the dissimilar objects are modeled by distant points with high probability.

\begin{figure}[t]
	\includegraphics[scale=0.50]{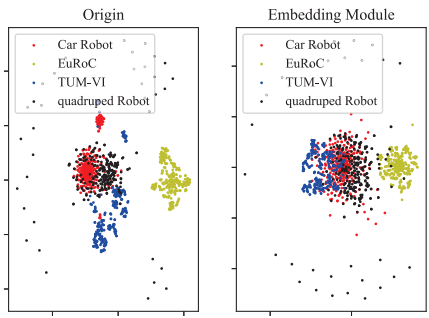}
	\centering
	\caption{Visualization of extracted representations of IMU sequence from different domains. Here, the EuRoC uses uncalibrated IMU.}
	\label{tsne}
\end{figure}

The ability of our Embedding module is qualified by the t-SNE projection (a tool to visualize high-dimension data) to show IMU sequences from multiple domains to an identical representation. The same t-SNE parameters (Perplexity=10, step=1000) are applied.  As can be seen in Fig. \ref{tsne}, originally, data points from different domains are distinctly separated into four folds. This proves that the datapoints from different tasks are dissimilar with each other. However, after the representation of Embedding module, the data points are scattered more dispersively. Even for EuRoC (the uncalibrated IMU), the distribution of its data points has intersections with that of others. This proves that the datapoints through Embedding module are nmore similar with each other.  This verifies the effectiveness of Embedding module and domain adaptability of our method.

\section{Conclusion}

 We  propose a few-shot domain adaptation method to improve the IMU adaptability in multiple scenarios.
To achieve the error of angular velocity, we propose a domain adaptation framework composed by Embedding module, Restructor module and Generator module.
The reconstitution loss is designed to improve domain adaptability.
In addition, we adopt a few-shot training strategy for further improving the adaptability in the case of limited data. 
In the experiment, we first test our method on two datasets (EuRoC and TUM-VI). Performances of the proposed framework and the proposed few-shot learning trategy are verified on the RSME and the whole process of the sequence. We also implement our methods on two real robots with three kinds of IMUs. Besides, t-SNE is used to visualize the results of Embedding module on datasets and real robots. This further proves the adaptability of our method.

\bibliographystyle{IEEEtran}
\bibliography{ref}

\end{document}